\documentclass[10pt,journal,compsoc]{IEEEtran}
\usepackage{cite}
\usepackage{amsmath,amssymb,amsfonts}
\usepackage{algorithmic}
\usepackage{graphicx}
\usepackage{textcomp}
\usepackage{multirow}
\usepackage{xcolor}
\usepackage{subfigure}
\usepackage{caption}
\usepackage{subfig}
\usepackage{float}

\if CLASSOPTIONcompsoc
  \usepackage[nocompress]{cite}
  \usepackage{cite}
\fi
\ifCLASSINFOpdf
\else
\fi
\begin{document}
%
\title{MoSFPAD: An end-to-end Ensemble of MobileNet and Support Vector Classifier for Fingerprint Presentation Attack Detection}
%
%
%
%

\author{Anuj Rai,~\IEEEmembership{}
        Somnath Dey,~\IEEEmembership{Senior Member IEEE}
        Pradeep Patidar,
        and~Prakhar Rai~\IEEEmembership{}
\IEEEcompsocitemizethanks{\IEEEcompsocthanksitem Anuj Rai, Somnath Dey, Pradeep Patidar,and Prakhar Rai are with the Department of Computer Science and Engineering, Indian Institute of Technology Indore, Indore, India, 453552.\protect\\
E-mail: phd1901201003@iiti.ac.in, somnathd@iiti.ac.in
\IEEEcompsocthanksitem}
\thanks{}}

%
%

\markboth{IEEE Transactions on Emerging Topics in Computing,~Vol.~XX, No.~X, XXXXX~XXXX}%
{Shell \MakeLowercase{\textit{et al.}}: }
%

\IEEEtitleabstractindextext{%
\begin{abstract}
Automatic fingerprint recognition systems are the most extensively used systems for person authentication although they are vulnerable to Presentation attacks. Artificial artifacts created with the help of various materials are used to deceive these systems causing a threat to the security of fingerprint-based applications. This paper proposes a novel end-to-end model to detect fingerprint Presentation attacks. The proposed model incorporates MobileNet as a feature extractor and a Support Vector Classifier as a classifier to detect presentation attacks in cross-material and cross-sensor paradigms. The feature extractor's parameters are learned with the loss generated by the support vector classifier. The proposed model eliminates the need for intermediary data preparation procedures, unlike other static hybrid architectures. The performance of the proposed model has been validated on benchmark LivDet 2011, 2013, 2015, 2017, and 2019 databases and overall accuracy of 98.64\%, 99.50\%, 97.23\%, 95.06\%, and 95.20\% are achieved on these databases, respectively. The performance of the proposed model is compared with state-of-the-art methods and the proposed method outperforms in cross-material and cross-sensor paradigms in terms of average classification error.
\end{abstract}

}
\maketitle

\IEEEdisplaynontitleabstractindextext

\IEEEpeerreviewmaketitle

\IEEEraisesectionheading{\section{Introduction}\label{sec:introduction}}
\IEEEPARstart{A}{utomatic} Fingerprint Recognition System (AFRS) is an easy, cost-effective, and user-friendly method of person authentication \cite{SS1}. Compared with other biometric systems, it requires less time, resources, and human effort to validate a person’s identity \cite{SS2}. Due to these reasons, it is being adapted by various commercial organizations and security agencies for person verification and authentication. However, AFRS are vulnerable to various challenges. One of them is a Presentation Attack (PA), which involves presenting a fabricated artifact of a genuine user's fingerprint to the sensor and gaining access to the system. Various fabrication materials such as woodglue, gelatine, modasil, and siligum are used to launch PAs on fingerprint biometric systems. Fingerprint Presentation Attack Detection (FPAD) is a countermeasure to deal with such types of attacks. FPAD methods are classified into two broad categories that are hardware-based methods and software-based methods. The hardware-based methods are costly due to the involvement of extra hardware devices that measures temperature, pulse, blood pressure, humidity, etc. On the other hand, software-based methods require only the fingerprint sample as input, making them cost-effective and user-friendly. Hence, our work is focused on the development of a software-based approach to detect PAs in AFRS. 

The state-of-the-art software-based approaches include perspiration and pore based-methods \cite{b27_derakshini, b16_espinoza, b19_abhyankar}, statistical and handcrafted feature based-methods \cite{xia_2,b13_xia1,b_Gragnaniello,b37_deepika,b5_sharma1,b1_dubey,b23_kim,b19_abhyankar}, and deep learning based-methods  \cite{b10_anusha,b26_arora,b14_chugh1,b15_chugh2,b4_uliyan, Nogueira,b47_jung2,Spinoulas}. Perspiration is a natural property that is seen in live fingers and not in spoofs. It gets affected by some external factors such as the temperature of the surroundings as well as the pressure applied on the fingertip. Due to the aforementioned reasons, sometimes multiple impressions of the same finger are required which causes inconvenience to the user. Pore-based methods rely on the quality of the input sample. Since a pore is a very small hole that is present in the live finger, sometimes it is hard to capture by the sensing devices. The statistical feature and handcrafted feature-based methods extract the predefined features and feed them to the classifier for classification. The use of different sensing devices results in considerable differences in the quality of fingerprint samples which has an impact on the performance of these approaches. However, some of these methods \cite{b5_sharma1,b37_deepika} have shown a remarkable performance in the intra-sensor paradigm but are not capable to detect the PAs in cross-sensor and cross-material paradigms. In recent years, researchers suggested some approaches which utilize Deep Convolutional Neural Networks (CNN) as a classifier. CNN's are proven to be a useful tool in the field of computer vision due to the possession of convolutional operations which is also capable of extracting minute features from the input samples. The literature suggested in \cite{b10_anusha,b14_chugh1,Nogueira} shows their superiority over the traditional methods in the detection of PAs but the detection of PAs in cross-material and cross-sensor paradigms is still a challenging issue.

In this paper, we propose a hybrid approach that incorporates both machine learning and deep learning models to develop an end-to-end architecture. MobileNet is a CNN architecture that uses depth-wise separable convolution instead of the traditional convolution operation. On the other side, SVC is proven to be an efficient classifier however it depends on feature extraction methods. In our work, we utilize MobileNet as a feature extractor and SVC as a classifier to develop an end-to-end architecture to deal with the PAs. This adaptation makes it suitable for devices with limited computational resources and increases the classification capability of MobileNet. The suggested model outperforms various existing methods of detecting PAs in cross-material and cross-sensor paradigms.
The main contributions of this paper are highlighted below.

1. A novel end-to-end ensemble of MobileNet and SVC is proposed which efficiently uses the characteristics of both deep CNN and SVC.

2. In contrast to the usual softmax or sigmoid function, the feature extractor learns its parameters using the loss generated by the SVC. The proposed architecture addresses the issues of developing an ensemble of CNN and SVC in a better way as it does not require any intermediary stage of data preparation for the SVC, unlike other hybrid designs. 

3. In the proposed approach, MobileNet is used as a feature extractor. As compared with other state-of-the-art CNNs, it reduces the demand for computational resources making it suitable for mobile devices.

4. The performance of the proposed model has been evaluated on a wide range of LivDet databases which is not reported in the existing single literature.

5. An exhaustive comparison of the proposed model has been done in intra-sensor, cross-material, and cross-sensor paradigms with state-of-the-art methods where the proposed model outperforms others.

The remainder of this paper is organized as follows. Section \ref{related work} presents the details of state-of-the-art methods suggested for the detection of PAs. Section \ref{proposed work} describes the working of the proposed architecture and in section \ref{experimental results}, experimental results, as well as comparative analysis, are given. The conclusion is discussed in section \ref{conclusion}.

\section{Related Work} \label{related work}
PAs are the most concerning security threat to the AFRS. In this section, several methods suggested by researchers to protect the AFRS from PAs are discussed. These methods can be further classified as per the methodology and utilization of resources. They are classified as perspiration and pore feature-based methods, statistical and handcrafted feature-based methods, and deep learning-based methods. This section discusses the approaches that fall under these categories along with their advantages and disadvantages.
\subsection{Perspiration and pore-based methods} Live fingers demonstrate a natural property of perspiration. The small holes or pores in them, cause sweat or perspiration due to the environmental temperature. Pores are small enough that they are hard to reflect in fabricated spoofs at the time of fabrication. This natural property of the live finger is exploited by researchers to discriminate between live and spoof samples. 
Derakshini et al. \cite{b27_derakshini} utilized the fingerprint sample's prevalence of perspiration and sweat diffusion patterns to distinguish between live and spoof fingerprints.
Espinoza et al. \cite{b16_espinoza} utilized the number of pores as a feature for the detection of PAs since a spoof sample does not contain the number of pores the same as in a live fingerprint sample. The proposed method is validated on a custom-made database. Further, this property is utilized by Abhyankar et al. \cite{b19_abhyankar}. They suggested a wavelet-based approach for the detection of PAs by exploiting the sweat feature of fingerprints. Although these methods are quite useful in the detection of PAs, they depend on the sensing device as well as the pressure of the fingertip for the extraction of the features. Sometimes, even a live fingerprint gets discarded by these FPAD system and requires multiple impressions of the same fingertip with a certain pressure. This makes perspiration and pore-based methods less user-friendly and tedious.

\subsection{Statistical and handcrafted feature-based methods}
Statistical and handcrafted feature-based methods extract quality-related information from the fingerprint samples for the discrimination of live and spoof samples. The skin of the fingers and the spoofing materials have different elasticity, moisture level, color, etc. which can be derived as features to detect PAs. In this section, we discuss the methods that utilize statistical or handcrafted features to extract the properties of the fingerprint for the detection of PAs.
Choi et al. \cite{b7_choi1} utilized statistical features i.e. histogram, directional contrast, ridge-thickness, and ridge-signal for the training of SVM. The proposed method is validated using a custom-made database. Similarly, Park et al. \cite{b41_park2} trained an SVM classifier for liveness detection on the ATVS Fake Fingerprint (ATVSFFp) database using statistical features such as deviation, variance, skewness, kurtosis, hyper skewness, and hyper flatness, as well as three additional features that are average brightness, standard deviation, and differential image. In \cite{b25_marasco}, Marasco et al. proposed a feature-based method that emphasized on statistical characteristics of a fingerprint sample. In this work, first-order statistical features and intensity-based features are extracted to train various classifiers as every image consists of texture which can provide information related to the grey level intensity and variation. This work is validated on LivDet 2009 database. Xia et al. \cite{b13_xia1} suggested a novel technique that extracts the second and third-order co-occurrence matrix of gradients and uses these as features to train the SVM classifier. Authors validated their method on LivDet 2009 and 2011. In another work, Xia et al. \cite{xia_2} developed an image descriptor that combines intensity variance and gradient-based properties to form a feature vector. This feature vector is further used to train the SVM classifier. The proposed method is tested on LivDet 2011 and 2013 databases.
Gragnaniello et al. \cite{b_Gragnaniello} suggested the use of a weber image descriptor to extract digital excitation and gradient information from fingerprint samples. Then the extracted features are fed to SVM for classification. This method is validated on the benchmark fingerprint databases. Since the live finger and its fabricated spoof have a different levels of elasticity, these features play a vital role in the detection of PAs. Sharma et al. \cite{b5_sharma1} utilized some quality-related handcrafted features such as Ridge and Valley Clarity (RVC), Ridge and Valley Smoothness (RVS), number of abnormal ridges and valleys, Orientation Certainty Level (OCL), Frequency Domain Analysis (FDA), etc.  The extracted features are combined to train the random forest classifier. The suggested model is tested on LivDet 2009, 2011, 2013, and 2015 databases.
Similarly, Dubey et al. \cite{b1_dubey} emphasized shape as well as texture-based features for the detection of PAs. In their work, Speeded Up Robust Feature (SURF) and Pyramid extension of Histogram of Gradients (PHOG) are utilized to extract the shape as a liveness property. In addition to these features, the Gabor wavelet is used to analyze the texture information. The method is validated using Livdet 2011 and 2013 databases. Kim et al. in \cite{b23_kim} suggested a unique image descriptor based on the local coherence of a fingerprint image. The texture of the same finger's fabricated samples differs significantly while being fabricated with different materials. This method leverages local coherence patterns as a feature to train the SVM and has been validated using LivDet 2009, 2011, 2013, and 2015 and ATVSFFp databases. Ghiani et al. \cite{b12_ghiani} used a local image descriptor known as Binary Statistical Image Feature (BSIF). This feature is produced by applying a series of natural filters on fingerprint samples. Then the results of these filters are translated into a binary sequence. The suggested technique has been evaluated on the Livdet 2011 database, but not in the cross-material and cross-sensor paradigms. Ajita et al. proposed an additive learning approach in \cite{b2_rattani1}. Unlike other methods, the suggested model classifies an input fingerprint sample into three categories: live, spoof, and unknown. The samples classified as `unknown' are utilized to train the model again. The proposed method is tested on a small set of fingerprint samples. In continuation to their previous work, Ajita et al. \cite{b3_rattani2} suggested a novel approach that utilizes a Weibull-calibrated Support Vector Machine (W-SVM) as a classifier. This SVM combines one-class and binary SVMs. This modification has shown to be a significant improvement over their previous efforts. Yuan et al. \cite{b24_yuan1} proposed an approach that depends on the generation of two co-occurrence matrices using the Laplacian operator. For various quantization operations, this operator is used to compute image gradient values. These gradient values are then used as a feature for the training of a backpropagation neural network. LivDet 2013 dataset was used to validate this technique. These methods rely on the quality of input fingerprint samples for the extraction of discriminating features. However, some of the aforementioned methods \cite{b5_sharma1, b37_deepika, b16_espinoza} have shown exceptional performance in the intra-sensor paradigm but do not exhibit the same performance in cross-sensor and cross-material scenarios.

\subsection{Deep learning based-methods}
Deep CNNs are proven to be great classifiers in the field of computer vision \cite{imagenet}. They consist of a set of convolutional filters that can extract discriminatory features from live and spoof images. In recent years, CNN-based PAD methods have been suggested by various researchers to confront various real-life object and image classification problems. By looking at their supremacy in image classification, researchers utilized them for the detection of PAs. Some of the suggested methods are discussed in this section.
Arora et al. \cite{b26_arora} presented a robust approach for detecting PAD that incorporates VGG architecture as a classifier. Fingerprint samples after contrast enhancement are fed to the VGG classifier for classification. The performance of the model is validated on various fingerprint databases, including FVC 2006, ATVSFFp, Finger vein dataset, LivDet 2013, and 2015 databases.
In \cite{Nogueira}, Nogueira et al. used pre-trained CNN architectures in their research. They employed VGG, Alexnet, CNN and SVC empowered with well-known hand-crafted features. The approach is validated using Livdet 2009, 2011, and 2013 databases. 
In \cite{b4_uliyan}, Uliyan et al. suggested the Deep Boltzmann Machine (DBM) and Restricted Boltzmann Machine (RBM) for extracting and determining a relationship between features. DBMs are proven to be useful in extracting features from fingerprint images. These methods outperform various handcrafted feature-based methods but lack in terms of performance against fingerprints created with unknown spoofing materials.
The aforementioned methods have not been tested in cross-material and cross-sensor scenarios and are not suitable for mobile devices due to the architecture adopted by them. The method suggested by \cite{Nogueira} utilizes multiple architectures for the classification which in turn, requires more computational resources. Anusha et al. in \cite{b10_anusha} proposed a multi-modal CNN-based FPAD method. It utilizes channel attention and spatial attention modules along with the proposed patch attention module. Two DenseNet classifiers with different configurations work together to produce liveness scores which are fused to get a global liveness score. This method is validated on LivDet 2011, 2013, 2015, and 2017 databases. The suggested model imposes heavy computational requirements due to the use of two DenseNet architectures in parallel. 
Further, Chugh et al. in \cite{b15_chugh2} proposed an approach that divides a fingerprint sample into patches based on the minutiae points present in it. The liveness score is predicted by the model for all the patches which are fused together in order to generate the global liveness score. This approach has been tested on the Livdet 2013 and 2015 and 2017, as well as on the MSU-FPAD dataset.
In continuation to their previous work, Chugh et al.\cite{b14_chugh1} proposed a novel approach for detecting PAs in cross-material and cross-sensor paradigms. They introduced a synthetic fingerprint generator by modifying an existing VGG-16 CNN architecture. This wrapper generates new fingerprint patches by image synthesis, which are used along with training data in order to train the classifier. This method is tested on LivDet 2015, and 2017 databases. The classification time of a single fingerprint sample is reported to be 100 milli-seconds on highly configured processors but the same may be computationally heavy for the device with limited computational resources due to the division of a single sample into around 45-50 samples.
Deep learning-based methods have shown remarkable performance while the detection of PAs but have imposed the need for more computational resources due to the involvement of multiple layers. In our work, we have utilized a lightweight deep CNN architecture with SVC in an end-to-end manner for better classification performance in intra-sensor, cross-material, and cross-sensor paradigms of FPAD.

\section{Proposed Work} \label{proposed work}
An end-to-end architecture with MobileNet and SVC for Fingerprint Presentation Attack Detection (MoSFPAD) is proposed in this work. MobileNet extracts the minute features from input fingerprint samples while SVC is utilized for the classification using those features. The proposed architecture also addresses the issue of developing a hybrid model which includes CNN and SVC. Further, the use of MobileNet reduces the need for computational resources and makes our architecture faster and more suitable for devices with limited computational resources. In the following subsections, a description of the MobileNet, Support Vector Classifier, and the proposed MoSFPAD is provided.

\subsection{\textbf{ MobileNet V1}} 
\begin{figure*}[h]
	\centering
	\resizebox{0.40\textheight}{!}{
		{
			
			\includegraphics[]{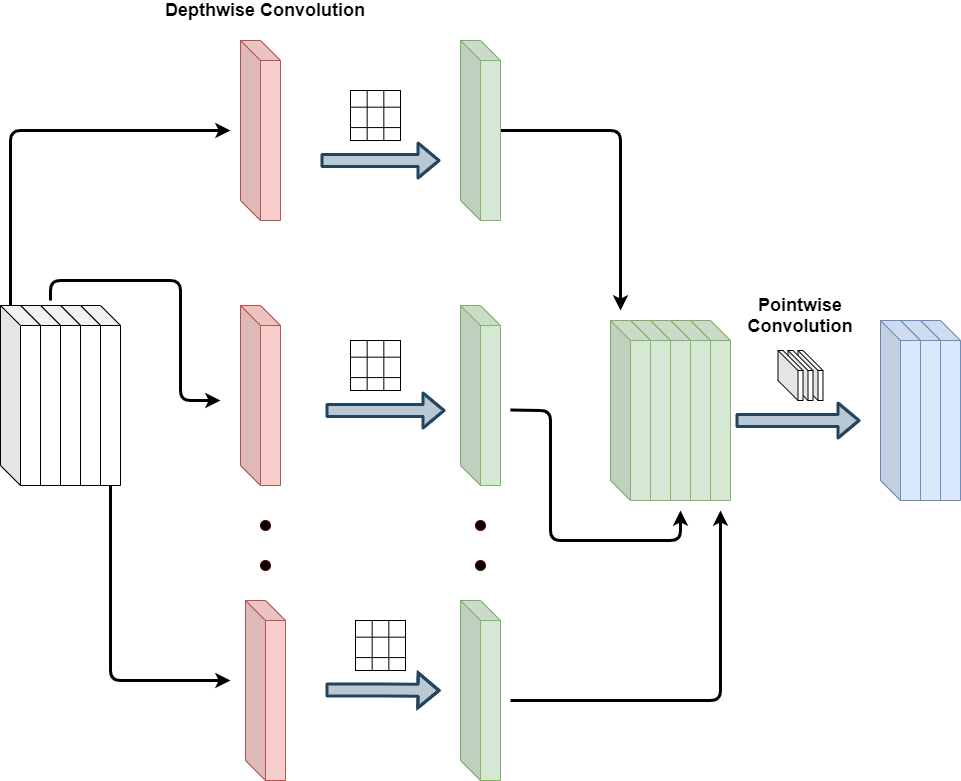}
	}
	}
	\caption{Depth-wise separable convolution.}
\vspace{3mm}	
\label{Hybrid_1}
\end{figure*}

We have opted MobileNet \cite{MobileNet} as a feature extractor to develop an FPAD model for devices with limited computational resources. It is advantageous over state-of-the-art CNN architectures as it utilizes depth-wise separable convolution instead of standard convolution. A standard convolution operation takes input as $X$ channels of size $D_{x} \times D_{x}$ and produces $D_{y} \times D_{y} \times Y$ feature maps by applying $D_{k} \times D_{k} \times Y$ filters where the spatial height and width of the squared input image are denoted with $D_{x}$. $X$ denotes the number of input channels while $D_{y}$ is the spatial height and width. The number of output feature maps is denoted with $Y$. Equation (\ref{standard conv}) describes the calculation of the output feature map for standard convolution operation\cite{convolution} having stride one with padding. \\

\begin{equation}
\label{standard conv}
G_{k,s,y}  = \sum_{p,q,x} (K_{p,q,x,y} \times F_{k+p-1,s+q-1,x}) 
\end{equation}
The computation cost of the convolution operation is given as $(D_{k} \times D_{k} \times X \times Y \times D_{y} \times D_{y}$). This procedure necessitates a large number of calculations making it unsuitable for devices with fewer computational resources. The MobileNet \cite{MobileNet}, adapts the depth-wise separable convolution which constitutes the convolution operation into depth-wise convolution followed by point-wise convolution as per Fig. \ref{Hybrid_1}. This tweak reduces the amount of computation to a large extent. Equation (\ref{depthwise conv}) denotes the depth-wise convolution operation.
\begin{equation}
\label{depthwise conv}
\hat{G}_{k,s,x}  = \sum_{p,q}(\hat{K}_{p,q,x} \times F_{k+p-1,l+q-1,x})
\end{equation}
where \^K is the depth-wise convolutional kernel of size $D_{k} \times D_{k} \times X$. In this case, the $x^{th}$ filter is applied to the $x^{th}$ channel in $F$ to generate the $x^{th}$ channel of the filtered output feature map $G$. The depth-wise separable convolution operation's computing cost is represented by ($D_{k} \times D_{k} \times X \times D_{y} \times D_{y}$).
Further, the point-wise convolution filter is applied to the generated feature maps. The cost of the point-wise convolution operation is denoted as ($X \times Y \times D_{y} \times D_{y}$). 
The overall cost of the depth-wise separable convolution operation is denoted by Eq. (\ref{depthwise total cost}).
\begin{equation}
\label{depthwise total cost}
Cost = D_{k} \times D_{k} \times X \times D_{y} \times D_{y} + X \times Y \times D_{y} \times D_{y} 
\end{equation}
The speedup ratio (S) of the depth-wise separable convolution over traditional convolution operation is denoted as Eq. (\ref{speedup}).
\begin{equation}
\label{speedup}
S = \frac{D_{k} \times D_{k}\times  X\times Y\times D_{y}\times D_{y}}{D_{k}\times D_{k}\times X\times D_{y}\times D_{y} + X\times Y\times D_{y}\times D_{y}}
\end{equation}
On solving Eq. (\ref{speedup}), we get the speedup (S), as $D^{2}_{k} + Y$ which clearly indicates the supremacy of the depth-wise separable convolution over traditional convolution operation in terms of computational cost. The utilization of depth-wise convolution makes it a low-latency network. This MobileNet has 1000 neurons in the last layer to classify the images belonging to 1000 classes. We have modified it for the binary classification problem. However, the MobileNet is cost-effective the adaptation of the depth-wise convolution operation costs them in terms of classification accuracy. In this work, we have suggested a novel ensemble technique to deal with the aforementioned issue by combining the MobileNet with SVC in a dynamic manner. 

\subsection{\textbf{Support Vector Classifier (SVC)}}
SVC is a supervised learning algorithm that is used for solving classification problems. The main objective of an SVC is to find a hyperplane that separates the data points in a high-dimensional space. SVC learns its parameters W by solving an optimization problem. The loss function of the SVC is denoted by Eq. (\ref{SVMEq}).
\begin{equation}
\label{SVMEq}
Loss = C\sum_{i=1}^{p} (max(0, 1 - y_{i}^{'}(W^{T}X_{i}+b)))
\end{equation}
Here, $W$ is the weight, $b$ is the bias and $X$ is the input feature vector having $p$ samples. 
The results reported in \cite{b3_rattani2, b_Gragnaniello, b7_choi1, b13_xia1, xia_2} indicate that SVM is a great classifier while being applied on appropriate features. After going through the advantages mentioned in the literature we decided to opt for SVC as a classifier in our work.
\subsection{\textbf{Proposed MoSFPAD }} End-to-end (E2E) learning is considered as training of a potentially complicated learning system using a single model (particularly a Convolutional Neural Network) that represents the designated system. In this paper, we propose a model that combines MobileNet with SVC for the detection of PAs in AFRS. Unlike other hybrid models \cite{daksha,meenakshi1,meenakshi2}, it is an end-to-end architecture in which the first module which is a feature extractor is trained with the loss generated by the SVC. As a feature extractor, MobileNet is found to be best suitable as it utilizes depth-wise separable convolution operation \cite{depthwise} instead of traditional convolution. The original MobileNet uses BinaryCrossEntropy as a loss function which is replaced with hinge loss since the last layer has been configured as an SVC.
\begin{figure*}[h]
	\centering
	\resizebox{0.5\textheight}{!}{
		{
			
			\includegraphics[]{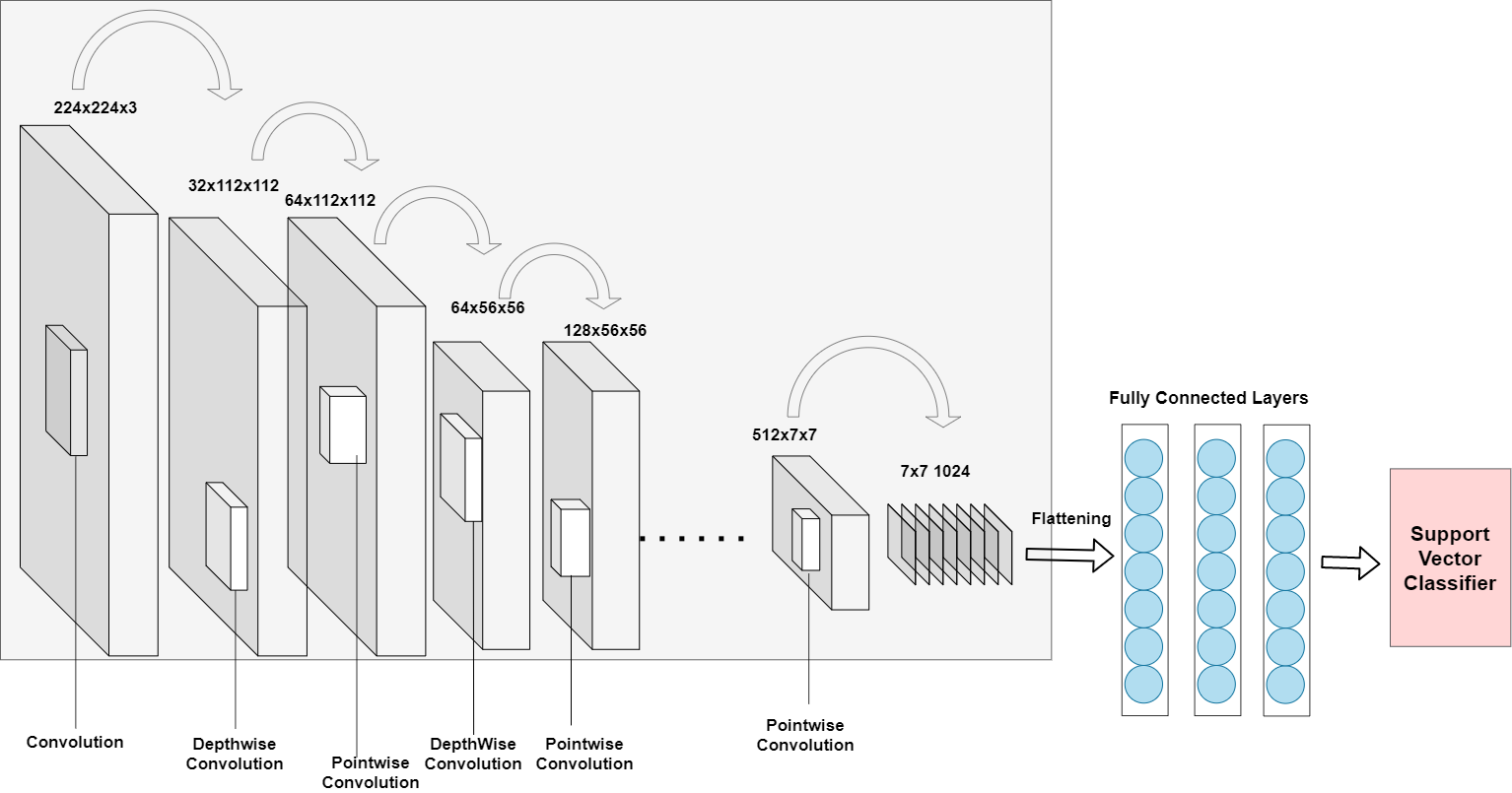}
	}}
	\caption{Block diagram of MoSFPAD architecture}
\vspace{3mm}	
\label{Hybrid_3}
\end{figure*}
The development of the proposed MoSFPAD architecture is explained in the following subsections.
\subsubsection{\textbf{Designing MoSFPAD Architecture}}
Figure \ref{Hybrid_3} depicts the architecture of $MoSFPAD$ where MobileNet acts as a feature extractor and SVC as a classifier. In MobileNet, all the convolutional layers are followed by batchnormalization and ReLU non-linearity.  MobileNet has 5 logical blocks each of which reduces the size of feature maps by half by performing downsampling using the max-pooling operation. The convolutional layer at the last returns 1024 feature maps of size 7 x 7. After this, we added three layers and one redesigned layer which is configured to work like an SVC. The output of the last layer is denoted with Eq. (\ref{last layer output})
\begin{equation}
\label{last layer output}
Output =  (W)^{T} CNN_{out} +b
\end{equation}
Here, $CNN_{out}$ is the output from the CNN which is a feature set of 256 values, $W$ and $b$ are weights as well as bias of the last layer,  $y_{i}^{'}$ is the actual class label of the input sample that is either -1 or +1.
The proposed modification makes the CNN learn its parameters with the loss generated by the proposed layer due to which the loss function has been modified as per the hinge loss. Equation \ref{last layer} denotes the formulation of loss calculated for training the parameters of MobileNet. 
\begin{equation}
\label{last layer}
Loss = C\sum_{i=1}^{p} max(0, 1 - y_{i}^{'}((W)^{T} CNN_{out}+b))
\end{equation}
Since the model has a huge number of learnable parameters, we have utilized Adam as an optimizer for learning of parameters. Initialized learning rate and weight decay values are 0.0001 and 0.0004, respectively.
\subsubsection{\textbf{Data Augmentation}} The Deep CNN architecture requires huge data for their training. However, LivDet databases consist of a limited amount of fingerprint images. The higher dimension of the input image causes a higher number of learnable parameters. It also tends the deep CNN architecture to suffer from the problem of overfitting. Data augmentation techniques are adapted to prevent the proposed model from suffering from overfitting. For the same, training data undergoes augmentation with rotation, flipping, and shear operations to generate more samples while the testing dataset is kept as it is. \\
\subsubsection{\textbf{Data Labelling}}
In binary classification, the data points belonging to a class are assigned as either 0 or 1 respectively. Similarly, the SVC classifies the data points by finding a hyperplane between the data points belonging to different classes. 
\vspace{2 mm}
\begin{center}
$W^{T}X$  =$\begin{cases}\label{cases}
=0, & \text{if, } $the point is on the hyperplane$ \\
<0, & \text{if, } $the point belongs to first class$ \\
>0, & \text{if, } $the point belongs to second class$  
\end{cases}$
\end{center} 
\vspace{2 mm}
The SVC function requires the class labels to be either -1 or +1 to draw a plane.
In this work, the last layer is replaced with SVC, and hence, fingerprint samples are assigned with the labels -1 for the spoof and +1 for the live.

\subsubsection{\textbf{Training of MoSFPAD}}
The proposed end-to-end MoSFPAD architecture is trained from scratch on benchmark LivDet databases. The parameters are initialized using imagenet weights instead of random weights for faster training of the feature extractor module. The findings of the proposed model in intra-sensor, cross-material, and cross-sensor are discussed in section \ref{experimental results}. The model produces a confidence score as an outcome for an input sample which is either a positive or a negative value. The input sample is considered live if the confidence score is positive and, spoofed if the confidence score is negative. Since the outcome is a real number,  we have utilized the min-max normalization technique to normalize these values between 0 and 1, which is denoted by Eq. (\ref{min-max}). 
\begin{equation}
\label{min-max}
Normalized Score = \frac{Confidence score - Min Value}{Max Value - Min Value}
\end{equation}
Where $MinValue$ and $MaxValue$ are the minimum and maximum values for a set of confidence scores computed by the model for samples present in a fingerprint database.
These normalized values are used for the evaluation of the model using the Detection Error Trade-off (DET) Curve which is discussed in section \ref{evaluation in high-security}.

\section{Experimental Results} \label{experimental results}
\subsection{\textbf{Database and Performance Metrics}}
In our work, experiments are carried out on the liveness detection competition LivDet 2011, 2013, 2015, 2017, and 2019 databases to validate the performance of the proposed model. Each database is prepared with multiple sensing devices. The involvement of multiple devices causes differences in the resolution of the sample images. For training and testing, fingerprint samples are given in separate sets. The details of the utilized benchmark databases are mentioned in Table \ref{Database_details}.
\begin{table*}[!hbt]
\begin{center}
\caption{Details of the databases}
\label{Database_details}
\resizebox{0.8\textwidth}{!}{
\begin{tabular}{|l|l|c|c|l|}
\hline
\textbf{Database}                     & \textbf{Sensor}          & \textbf{Live} & \textbf{Spoof} & \textbf{Spoofing   Materials}                                                      \\ \hline
\multirow{4}{*}{\textbf{LivDet 2011}} & \textbf{Biometrika}      & 1000/1000     & 1000/1000      & \multirow{2}{*}{Ecoflex, Gelatine, Latex, Siligum, Woodglue}                       \\ \cline{2-4}
                                      & \textbf{Italdata}        & 1000/1000     & 1000/1000      &                                                                                    \\ \cline{2-5} 
                                      & \textbf{Digital Persona} & 1000/1000     & 1000/1000      & \multirow{2}{*}{Gelatine, Latex, Playdoh, Silicone, WoodGlue}                      \\ \cline{2-4}
                                      & \textbf{Sagem}           & 1000/1000     & 1000/1000      &                                                                                    \\ \hline
\multirow{2}{*}{\textbf{LivDet 2013}} & \textbf{Biometrika}      & 1000/1000     & 1000/1000      & \multirow{2}{*}{Ecoflex, Gelatine, Latex, Modsil, Woodglue}                        \\ \cline{2-4}
                                      & \textbf{Digital Persona} & 1000/1000     & 1000/1000      &                                                                                    \\ \hline
\multirow{4}{*}{\textbf{LivDet 2015}} & \textbf{Crossmatch}      & 1000/1000     & 1473/1448      & Body Double, Ecoflex, Playdoh, OOMOO, Gelatine                                     \\ \cline{2-5} 
                                      & \textbf{Digital Persona} & 1000/1000     & 1000/1500      & \multirow{3}{*}{Ecoflex, Latex, Gelatine, Woodglue, Liquid Ecoflex,   RTV}         \\ \cline{2-4}
                                      & \textbf{Greenbit}        & 1000/1000     & 1000/1500      &                                                                                    \\ \cline{2-4}
                                      & \textbf{Hi-Scan}         & 1000/1000     & 1000/1500      &                                                                                    \\ \hline
\multirow{3}{*}{\textbf{LivDet 2017}} & \textbf{Greenbit}        & 1000/1700     & 1200/2040      & \multirow{3}{*}{Body Double, Ecoflex, Woodglue, Gelatine, Latex,   Liquid Ecoflex} \\ \cline{2-4}
                                      & \textbf{Orcanthus}       & 1000/1700     & 1180/2018      &                                                                                    \\ \cline{2-4}
                                      & \textbf{Digital Persona} & 999/1692      & 1199/2028      &                                                                                    \\ \hline
\multirow{3}{*}{\textbf{LivDet 2019}} & \textbf{Greenbit}        & 1000/1020     & 1200/1224      &            Body Double, Ecoflex, Woodglue, Mix1, Mix2, Liquid Ecoflex                                                                        \\ \cline{2-5} 
                                      & \textbf{Orcanthus}       & 1000/990      & 1200/1088      &   Body Double, Ecoflex, Woodglue, Mix1, Mix2, Liquid Ecoflex                                                                                 \\ \cline{2-5} 
                                      & \textbf{Digital Persona} & 1000/1099     & 1000/1224      &     Ecoflex, Gelatine, Woodglue, Latex, Mix1, Mix2, Liquid Ecoflex                                                                               \\ \hline
\end{tabular}}
\end{center}
\end{table*}

The performance of the proposed model is measured using ISO/IEc IS 30107 criteria \cite{misc1}. The Attack Presentation Classification Error Rate (APCER) gives the percentage of misclassified spoof fingerprint samples, and its counterpart, the Bonafide Presentation Classification Error Rate (BPCER) gives the percentage of misclassified live fingerprint samples. The Average Classification Error (ACE) is the averaged sum of APCER and BPCER. It is used to evaluate the overall performance of the FPAD system. Equation (\ref{ACE}) represents the calculation of ACE.
\begin{equation}
\label{ACE}
ACE = \frac{APCER + BPCER}{2}
\end{equation}
The ACE is further utilized to derive the accuracy of the proposed model using Equation (\ref{Accuracy}).
\begin{equation}
\label{Accuracy}
Accuracy = 100 -  ACE
\end{equation}
\subsection{\textbf{Implementation Details}}
The proposed model is implemented using Tensorflow-Keras library. All training and testing have been done on NVIDIA TESLA P100 GPU. The proposed algorithm is implemented in Python and the models are implemented using the Tensorflow-Keras library. Each model has been trained for 300 epochs which took time between 5 to 6 hours to converge. 
\subsection{\textbf{ Experimental Results}}
The proposed model is validated in three different scenarios i.e. intra-sensor and known spoof material, intra-sensor and unknown spoof material, and cross-sensor. A detailed description of these scenarios is given in the following subsections.
\subsubsection{\textbf{Intra-Sensor and Known Spoof Material}}
In this setup, the training and testing samples are captured with the same sensing device, and the spoof samples belonging to both datasets are fabricated using the same fabrication materials. The LivDet 2011 and 2013 databases are used for this setup while LivDet 2015 partially follows this configuration. The findings of the proposed model in terms of ACE are reported in Table \ref{tab: 2011_2013 intra-sensor} for LivDet 2011 and 2013 databases. Table \ref{tab: 2011_2013 intra-sensor} indicates that the proposed model achieves an average BPCER of 3.12\%, APCER of 1.21\%, and ACE of 2.15\% while being tested on LivDet 2011 database. Similarly, it attains an average BPCER of 0.22\%, APCER of 0.22\%, and the same for the ACE on LivDet 2013 database. The findings on LivDet 2015, which has spoof samples created using known and unknown materials, are reported in Table \ref{tab : 2015_Intra_Sensor}. It shows that the proposed model is able to detect the live samples with an average BPCER of 3.0\% only. Similarly, it detects the spoof samples created with known materials with an average APCER of 1.93\% as mentioned by the column ``APCER (Known)".
\begin{table}[!htb]
\caption{Intra Sensor performance on LivDet 2011, 2013 database}
\label{tab: 2011_2013 intra-sensor}
\begin{tabular}{|l|l|c|c|c|}
\hline
\textbf{Database} & \textbf{Sensor} & \textbf{BPCER} & \textbf{APCER} & \textbf{ACE (\%)} \\ \hline
\multirow{5}{*}{\textbf{LivDet 2011}} & \textbf{Biometrika} & 1.4 & 1.3 & 1.36 \\ \cline{2-5} 
 & \textbf{Digital Persona} & 0.41 & 0.30 & 0.36 \\ \cline{2-5} 
 & \textbf{Italdata} & 9.65 & 3.05 & 6.28 \\ \cline{2-5} 
 & \textbf{Sagem} & 1.02 & 0.20 & 0.60 \\ \cline{2-5} 
 & \textbf{Average} & \textbf{3.12} & \textbf{1.21} & \textbf{2.15} \\ \hline
\multirow{3}{*}{\textbf{LivDet 2013}} & \textbf{Biometrika} & 0.35 & 0.05 & 0.20 \\ \cline{2-5} 
 & \textbf{Italdata} & 0.10 & 0.40 & 0.25 \\ \cline{2-5} 
 & \textbf{Average} & \textbf{0.22} & \textbf{0.22} & \textbf{0.22} \\ \hline
\end{tabular}
\end{table}

\begin{table}[!bht]
\caption{Intra Sensor performance on LivDet 2015 database}
\label{tab : 2015_Intra_Sensor}
\resizebox{0.50\textwidth}{!}{
\begin{tabular}{|l|l|c|c|c|c|}
\hline
\textbf{Database} & \textbf{Sensor} & \textbf{BPCER} & \textbf{\begin{tabular}[c]{@{}l@{}}APCER\\ (Known)\end{tabular}} & \textbf{\begin{tabular}[c]{@{}l@{}}APCER \\ (Unknown)\end{tabular}} & \textbf{ACE (\%)} \\ \hline
\multirow{5}{*}{\textbf{LivDet 2015}} & \textbf{Crossmatch} & 1.69 & 1.08 & 2.96 & 1.61 \\ \cline{2-6} 
 & \textbf{Digital Persona} & 7.32 & 1.87 & 1.96 & 3.65 \\ \cline{2-6} 
 & \textbf{Biometrika} & 2.34 & 4.37 & 2.80 & 3.23 \\ \cline{2-6} 
 & \textbf{Greenbit} & 2.04 & 0.41 & 1.67 & 1.32 \\ \cline{2-6} 
 & \textbf{Average} & \textbf{3.0} & \textbf{1.93} & \textbf{2.34} & \textbf{2.45} \\ \hline
\end{tabular}}
\end{table}

\subsubsection{\textbf{Intra-Sensor and Unknown Spoof Material}}
In this setup, both the training and testing fingerprint samples are captured using the same sensing device while the fabrication materials utilized for the creation of spoof samples are different. The performance of an FPAD model is desired to be higher in this setup because it ensures its capability to detect the PAs in real-life scenarios where the intruder can devise novel material to fabricate more realistic spoofs of a genuine user's finger. LivDet 2017 and 2019 are collected in the same way and the ACE for these databases are reported in Table \ref{tab: intra_2017_2019}. Table \ref{tab: intra_2017_2019} shows that the proposed model achieves an average BPCER of 6.15\%, APCER of 4.16\%, and ACE of 4.95\% on LivDet 2017 database. Similarly, it classifies the live and spoof samples of LivDet 2019 database with an average BPCER of 5.55\%, and APCER of 4.15\%, respectively. The Average ACE is reported to be 4.79\% for the same. Since one-third of the testing spoof samples in the LivDet 2015 database are fabricated using unknown materials, LivDet 2015 partially belongs to this category. The column ``APCER (Unknown)" in Table \ref{tab : 2015_Intra_Sensor} shows that the proposed model confronts the spoofs created with unknown material and only 2.34\% of the spoofs samples deceive the system as reported in the column ``APCER (Unknown) in the same table.
\begin{table}[!bht]
\caption{Intra-Sensor performance on LivDet 2017, and 2019 databases}
\label{tab: intra_2017_2019}
\begin{tabular}{|l|l|c|c|c|}
\hline
\textbf{Database} & \textbf{Sensor} & \textbf{BPCER} & \textbf{APCER} & \textbf{ACE (\%)} \\ \hline
\multirow{4}{*}{\textbf{LivDet 2017}} & \textbf{Digital Persona} & 5.72 & 3.72 & 4.60 \\ \cline{2-5} 
 & \textbf{Orcanthus} & 7.32 & 5.75 & 6.19 \\ \cline{2-5} 
 & \textbf{Greenbit} & 5.42 & 3.03 & 4.06 \\ \cline{2-5} 
 & \textbf{Average} & \textbf{6.15} & \textbf{4.16} & \textbf{4.95} \\ \hline
\multirow{4}{*}{\textbf{LivDet 2019}} & \textbf{Digital Persona} & 9.63 & 9.23 & 9.41 \\ \cline{2-5} 
 & \textbf{Greenbit} & 4.27 & 1.39 & 2.69 \\ \cline{2-5} 
 & \textbf{Orcanthus} & 2.75 & 1.85 & 2.28 \\ \cline{2-5} 
 & \textbf{Average} & \textbf{5.55} & \textbf{4.15} & \textbf{4.79} \\ \hline
\end{tabular}
\end{table}
\subsubsection{\textbf{Cross-Sensor Validation}}
In this experimental setup, the training and testing samples captured using different sensing devices are used to assess the performance of the FPAD model. Similar to testing in the cross-material paradigm, testing in this paradigm is required for an FPAD method to be used in a real-world scenario. It ensures its platform in-dependency against different sensing devices utilized by different organizations to capture the fingerprint sample. The results in this setup on LivDet 2011, 2013, 2015, 2017, and 2019 are reported in Table \ref{cross-sensor_all}. Table \ref{cross-sensor_all} indicates that the proposed model achieves an ACE of 37.88\%, 3.31\%, and 30.48\% while being tested on LivDet 2011, 2013, and 2015 respectively. Similarly, it attains the ACE of 31.61\% and 40.93\% on LivDet 2015 and 2019 databases in the same testing scenario.

\begin{table*}[!bht]
\caption{Cross-sensor performance of the proposed method on LivDet 2011, 2013, 2015, 2017, and 2019 databases}
\label{cross-sensor_all}
\begin{center}

\begin{tabular}{|l|l|l|l|c|l|l|l|}
\hline
\textbf{Database}                      & \textbf{Training}                                                                     & \multicolumn{1}{c|}{\textbf{Testing}}                                                      & \multicolumn{1}{c|}{\textbf{\begin{tabular}[c]{@{}c@{}}ACE\\ (\%)\end{tabular}}} & \textbf{Database}                       & \multicolumn{1}{c|}{\textbf{Training}}                                                & \textbf{Testing}                                                                           & \multicolumn{1}{c|}{\textbf{\begin{tabular}[c]{@{}c@{}}ACE\\ (\%)\end{tabular}}} \\ \hline
                                       &                                                                                       & \textbf{\begin{tabular}[c]{@{}l@{}}Digital \\ Persona\end{tabular}}                        & \multicolumn{1}{c|}{27.9}                                                        &                                         &                                                                                       & \textbf{Greenbit}                                                                          & 40.83                                                                            \\ \cline{3-4} \cline{7-8} 
                                       &                                                                                       & \textbf{Italdata}                                                                          & \multicolumn{1}{c|}{42.55}                                                       &                                         &                                                                                       & \textbf{\begin{tabular}[c]{@{}l@{}}Digital \\ Persona\end{tabular}}                        & 47.62                                                                            \\ \cline{3-4} \cline{7-8} 
                                       & \multirow{-3}{*}{\textbf{Biometrika}}                                                 & \textbf{Sagem}                                                                             & \multicolumn{1}{c|}{23.73}                                                       &                                         & \multirow{-3}{*}{\textbf{Crossmatch}}                                                 & \textbf{Biometrika}                                                                        & 42.05                                                                            \\ \cline{2-4} \cline{6-8} 
                                       &                                                                                       & \textbf{Biometrika}                                                                        & 44.06                                                                            &                                         &                                                                                       & \textbf{Crossmatch}                                                                        & 28.77                                                                            \\ \cline{3-4} \cline{7-8} 
                                       &                                                                                       & \textbf{Italdata}                                                                          & 42.51                                                                            &                                         &                                                                                       & \textbf{\begin{tabular}[c]{@{}l@{}}Digital \\ Persona\end{tabular}}                        & 12.53                                                                            \\ \cline{3-4} \cline{7-8} 
                                       & \multirow{-3}{*}{\textbf{\begin{tabular}[c]{@{}l@{}}Digital \\ Persona\end{tabular}}} & \textbf{Sagem}                                                                             & 44.80                                                                            &                                         & \multirow{-3}{*}{\textbf{Greenbit}}                                                   & \textbf{Biometrika}                                                                        & 29.56                                                                            \\ \cline{2-4} \cline{6-8} 
                                       &                                                                                       & \textbf{Biometrika}                                                                        & \multicolumn{1}{c|}{22.25}                                                       &                                         &                                                                                       & \textbf{Crossmatch}                                                                        & 29.86                                                                            \\ \cline{3-4} \cline{7-8} 
                                       &                                                                                       & \textbf{\begin{tabular}[c]{@{}l@{}}Digital \\ Persona\end{tabular}}                        & \multicolumn{1}{c|}{34.25}                                                       &                                         &                                                                                       & \textbf{Greenbit}                                                                          & 25.11                                                                            \\ \cline{3-4} \cline{7-8} 
                                       & \multirow{-3}{*}{\textbf{Italdata}}                                                   & \textbf{Sagem}                                                                             & \multicolumn{1}{c|}{34.09}                                                       &                                         & \multirow{-3}{*}{\textbf{\begin{tabular}[c]{@{}l@{}}Digital\\ Persona\end{tabular}}}  & \textbf{Biometrika}                                                                        & 39.88                                                                            \\ \cline{2-4} \cline{6-8} 
                                       &                                                                                       & \textbf{Biometrika}                                                                        & 38.55                                                                            &                                         &                                                                                       & \textbf{Crossmatch}                                                                        & 36.62                                                                            \\ \cline{3-4} \cline{7-8} 
                                       &                                                                                       & \textbf{\begin{tabular}[c]{@{}l@{}}Digital \\ Persona\end{tabular}}                        & 46.86                                                                            &                                         &                                                                                       & \textbf{Greenbit}                                                                          & 14.68                                                                            \\ \cline{3-4} \cline{7-8} 
                                       & \multirow{-3}{*}{\textbf{Sagem}}                                                      & \textbf{Italdata}                                                                          & 50.05                                                                            &                                         & \multirow{-3}{*}{\textbf{Biometrika}}                                                 & \textbf{\begin{tabular}[c]{@{}l@{}}Digital\\ Persona\end{tabular}}                         & 18.34                                                                            \\ \cline{2-4} \cline{6-8} 
\multirow{-13}{*}{\textbf{LivDet 2011}}         & \textbf{}                                                                             & \textbf{Average}                                                                           & \multicolumn{1}{c|}{\textbf{37.88}}                                              & \multirow{-13}{*}{\textbf{LivDet 2015}} & \textbf{Average}                                                                      & \textbf{}                                                                                  & \textbf{30.48}                                                                   \\ \hline
                                       &                                                                                       & \textbf{\begin{tabular}[c]{@{}l@{}}Digital \\ Persona\end{tabular}}                        & \multicolumn{1}{c|}{16.45}                                                       &                                         &                                                                                       & \textbf{\begin{tabular}[c]{@{}l@{}}Digital \\ Persona\end{tabular}}                        & 41.47                                                                            \\ \cline{3-4} \cline{7-8} 
                                       & \multirow{-2}{*}{\textbf{Greenbit}}                                                   & \textbf{Orcanthus}                                                                         & \multicolumn{1}{c|}{42.67}                                                       &                                         & \multirow{-2}{*}{\textbf{Greenbit}}                                                   & \multicolumn{1}{c|}{\textbf{Orcanthus}}                                                    & 45.63                                                                            \\ \cline{2-4} \cline{6-8} 
                                       &                                                                                       & {\color[HTML]{202124} \textbf{\begin{tabular}[c]{@{}l@{}}Digital \\ Persona\end{tabular}}} & \multicolumn{1}{c|}{38.71}                                                       &                                         &                                                                                       & {\color[HTML]{202124} \textbf{\begin{tabular}[c]{@{}l@{}}Digital \\ Persona\end{tabular}}} & 41.82                                                                            \\ \cline{3-4} \cline{7-8} 
                                       & \multirow{-2}{*}{\textbf{Orcanthus}}                                                  & \textbf{Greenbit}                                                                          & \multicolumn{1}{c|}{27.42}                                                       &                                         & \multirow{-2}{*}{\textbf{Orcanthus}}                                                  & \textbf{Greenbit}                                                                          & 31.87                                                                            \\ \cline{2-4} \cline{6-8} 
                                       &                                                                                       & \textbf{Greenbit}                                                                          & 22.61                                                                            &                                         &                                                                                       & \textbf{Greenbit}                                                                          & 39.41                                                                            \\ \cline{3-4} \cline{7-8} 
                                       & \multirow{-2}{*}{\textbf{\begin{tabular}[c]{@{}l@{}}Digital \\ Persona\end{tabular}}} & \textbf{Orcanthus}                                                                         & 41.75                                                                            &                                         & \multirow{-2}{*}{\textbf{\begin{tabular}[c]{@{}l@{}}Digital \\ Persona\end{tabular}}} & \textbf{Orcanthus}                                                                         & 45.39                                                                            \\ \cline{2-4} \cline{6-8} 
\multirow{-7}{*}{\textbf{LivDet 2017}}          & \textbf{}                                                                             & \textbf{Average}                                                                           & \textbf{31.61}                                                                   & \multirow{-7}{*}{\textbf{LivDet 2019}}  & \textbf{Average}                                                                      &                                                                                            & \textbf{40.93}                                                                   \\ \hline
                                       & \textbf{Biometrika}                                                                   & \textbf{Italdata}                                                                          & 5.93                                                                             & \multicolumn{1}{l|}{}                   &                                                                                       &                                                                                            &                                                                                  \\ \cline{2-8} 
                                       & \textbf{Italdata}                                                                     & \textbf{Biometrika}                                                                        & 0.7                                                                              & \multicolumn{1}{l|}{}                   &                                                                                       &                                                                                            &                                                                                  \\ \cline{2-8} 
\multirow{-3}{*}{\textbf{LivDet 2013}} & \textbf{}                                                                      & \textbf{Average}                                                                           & \textbf{3.31}                                                                    & \multicolumn{1}{l|}{}                   &                                                                                       &                                                                                            &                                                                                  \\ \hline
\end{tabular}
\end{center}
\end{table*}

\subsubsection{\textbf{Discussion}}
The live and spoof fingerprint samples have different textures, and ridge valley widths due to different elasticity levels of the finger skin and spoofing materials. The possession of convolutional layers enables the CNNs to classify the input samples by extracting the discriminating features from input fingerprint samples. On the other hand, the dynamic ensemble of SVC enables CNN to learn its parameters in a better way since the SVC tries to find a hyperplane between the samples of both classes. The findings of the proposed method are compared with existing methods tested on benchmark databases which are mentioned in the subsections below.
\subsubsection{\textbf{Comparison with Existing Approaches in Intra-Sensor Paradigm}}
The outcomes of the proposed method are compared to various software-based state-of-the-art techniques that are tested in this paradigm. Table \ref{comp_2011} indicates that the proposed method outperforms the methods discussed in \cite{xia_2}, \cite{b1_dubey}, \cite{b24_yuan1}, \cite{b_Gragnaniello}, \cite{Nogueira}, \cite{b24_yuan1}, \cite{b40_jian} over LivDet 2011 database. However, the performance of the proposed model is lower than the method discussed in \cite{b15_chugh2} on italdata sensor, but it is better on biometrika, digital persona, and sagem sensors. The overall accuracy of 97.89\% signifies that the proposed method has the capability to perform better while the spoofs are created using the cooperative method of spoofing.
\begin{table*}[!bht]
\begin{center}
\caption{Comparison with state-of-the-art methods on LivDet 2011 in intra-sensor paradigm } 
\label{comp_2011}
\resizebox{0.6\textwidth}{!}{
\begin{tabular}{|p{3.0 cm}|c|c|c|c|c|}
\hline
\textbf{Method} & \textbf{\begin{tabular}[c]{@{}c@{}}Accuracy\\(Biometrika)\end{tabular}}& \textbf{\begin{tabular}[c]{@{}c@{}}Accuracy\\(Digital Persona)\end{tabular}}& \textbf{\begin{tabular}[c]{@{}c@{}}Accuracy\\(Italdata)\end{tabular}}& \textbf{\begin{tabular}[c]{@{}c@{}}Accuracy\\(Sagem)\end{tabular}}& \multicolumn{1}{l|}{\textbf{Average}} \\ \hline
Xia et al. \cite{b13_xia1}           & 93.55                                   & 96.2                                          & 88.25                                  & 96.66                               & 93.37                                  \\ \hline
 Dubey et al. \cite{b1_dubey}         & 92.11                                   & 93.75                                         & 91.9                                   & 94.64                               & 93.1                                   \\ \hline
Yuan et al. \cite{b24_yuan1}          & 97.05                                   & 88.94                                         & 97.8                                   & 92.01                               & 93.82                                  \\ \hline
Gragnaniello et al. \cite{b_Gragnaniello}  & 93.1                                    & 92.00                                         & 87.35                                  & 96.35                               & 92.2                                   \\ \hline
Nogueira et al. \cite{Nogueira}      & 91.8                                    & 98.1                                          & 94.91                                  & 95.36                               & 95.04                                  \\ \hline
Yuan et al. \cite{b24_yuan1}          & 90.08                                   & 98.65                                         & 87.65                                  & 97.1                                & 93.55                                  \\ \hline
 Jian et al. \cite{b40_jian}          & 95.75                                   & 98.4                                          & 94.1                                   & 96.83                               & 96.27                                  \\ \hline
Chugh et al. \cite{b15_chugh2}                     & 98.76                                   & 98.39                                         & 97.55                                  & 98.61                               & 98.33                                  \\ \hline Sharma et al. \cite{b5_sharma1}                     &  92.7                                  &     94.4                                     &       88.6                           &    93.3                            &             92.25                      \\ \hline

                                       \textbf{MoSFPAD}          & \textbf{98.80}                          & \textbf{99.64}                                & \textbf{93.72}                         & \textbf{99.40}                      & \textbf{97.89}                         \\\hline
\end{tabular}}
\end{center}
\end{table*}

Similarly, a comparison is shown in Table \ref{comp_2013} describes that the proposed method achieves better classification accuracy than the methods discussed in \cite{b24_yuan1}, \cite{b40_jian}, \cite{b31_zhang}, \cite{b41_park2}, \cite{b42_Gottschlich}, \cite{b43_johnson}, \cite{b24_yuan1}, \cite{b45_jung}, \cite{b4_uliyan}, \cite{Nogueira}, \cite{b10_anusha}, and \cite{b14_chugh1} on LivDet 2013 database. The overall classification accuracy of 99.78\% indicates that the proposed method is able to perform better while the spoofs are created with the non-cooperative method of spoofing. 

\begin{table}[!bht]

\begin{center}
\caption{Comparison with state-of-the-art methods on LivDet 2013 in intra-sensor paradigm }    
\label{comp_2013}
\begin{tabular}{ |p{2.5 cm}| c| c| c|}
\hline
 \textbf{Method}                                                                                                    & \textbf{\begin{tabular}[c]{@{}c@{}}Accuracy\\(Biometrika)\end{tabular}} & \textbf{\begin{tabular}[c]{@{}c@{}}Accuracy\\(Italdata)\end{tabular}} & \multicolumn{1}{l|}{\textbf{Average}} \\ \hline
Yuan et al. \cite{b24_yuan1}                                                                                                       & 96.45                                             & 97.65                                           & 97.05                                 \\ \hline
 Jian et al. \cite{b40_jian}                                                                                                           & 99.25                                             & 99.40                                           & 99.32                                 \\ \hline
Zhang et al. \cite{b31_zhang}                                                                                                   & 99.53                                             & 96.99                                           & 98.26                                 \\ \hline
Park et al. \cite{b41_park2}                                                                                          & 99.15                                             & 98.75                                           & 98.95                                 \\ \hline
Gottschlich et al. \cite{b42_Gottschlich}                      & 96.10                                             & 98.30                                           & 97.0                                  \\ \hline
Johnson et al. \cite{b43_johnson}                                                                  & 98.0                                              & 98.4                                            & 98.20                                  \\ \hline
Yuan et al. \cite{b44_yuan3}                                                                            & 95.65                                             & 98.6                                            & 97.12                                 \\ \hline
Jung et al. \cite{b45_jung}                                                                                  & 94.12                                             & 97.92                                           & 96.02                                 \\ \hline
Uliyan et al. \cite{b4_uliyan}                                                                         & 96.0                                              & 94.50                                           & 95.25                                 \\ \hline
Nogueira et al. \cite{Nogueira}                                                                                  & 99.20                                             & 97.7                                            & 98.45                                 \\ \hline
Chugh et al. \cite{b15_chugh2}                                                                                  & 99.80                                             & 99.70                                            & 99.75                                 \\ \hline
Anusha et al. \cite{b10_anusha}                                                                                  & 99.76                                             & 99.68                                            & 99.72                                 \\ \hline
\textbf{MoSFPAD}                                                                                           & \textbf{99.80}                                    & \textbf{99.75}                                  & \textbf{99.78}                       \\ \hline
\end{tabular}
\end{center}
\end{table}

The performance of the proposed method is also compared with state-of-the-art methods tested on LivDet 2015 database which is mentioned in Table \ref{tab: Comparison_Intra_2015}. As per the comparison mentioned in Table \ref{tab: Comparison_Intra_2015}, it is evident that the proposed method performs well as compared with the method discussed in \cite{b41_park2}, \cite{b31_zhang}, \cite{b5_sharma1}, \cite{b46_zhang2}, \cite{Spinoulas}, \cite{b45_jung}, \cite{b34_livdet2015}, \cite{b4_uliyan}, and \cite{b23_kim} on LivDet 2015 database. It performs better than \cite{b31_zhang} on the dataset collected with crossmatch and biometrika sensors however on greenbit and digital persona, the performance is comparable. The proposed method also performs better than the method suggested by Jian et al. \cite{b40_jian} on crossmatch dataset while it is comparable on greenbit, digital persona, and biometrika datasets.

\begin{table*}[!hbt]
\begin{center}
 \caption{Comparison with state-of-the-art methods on LivDet 2015 in intra-sensor paradigm }
\label{tab: Comparison_Intra_2015} 
\resizebox{0.60\textwidth}{!}{
\begin{tabular}{|p{2.5cm}|c|c|c|c|c|}
\hline
 \textbf{Method}                    & \textbf{\begin{tabular}[c]{@{}c@{}}Accuracy\\(Crossmatch)\end{tabular}} & \textbf{\begin{tabular}[c]{@{}c@{}}Accuracy \\(Greenbit)\end{tabular}} & \textbf{\begin{tabular}[c]{@{}c@{}}Accuracy\\(Digital Persona)\end{tabular}} & \textbf{\begin{tabular}[c]{@{}c@{}}Accuracy\\ (Biometrika)\end{tabular}} & \textbf{Average}                      \\ \hline
                                     Park et al.\cite{b41_park2}                                                                                     & 99.63                                                                    & 97.30                                                                   & 91.5                                                                         & 95.9                                                                       & 96.08                        \\ \hline
                                    Jian et al. \cite{b40_jian}                                                                                      & 98.28                                                                    & 99.52                                                                   & 97.58                                                                        & 98.18                                                                      & 98.39                        \\ \hline
Zhang et al. \cite{b46_zhang2}                                                                                     & 97.05                                                                    & 99.47                                                                   & 96.39                                                                        & 97.05                                                                      & 97.78                        \\ \hline
Sharma et al. \cite{b5_sharma1}                                                                                    & 98.07                                                                    & 95.7                                                                    & 94.16                                                                        & 95.22                                                                      & 95.78                        \\ \hline
\begin{tabular}[c]{@{}l@{}}Spinoulas et al. \\ \cite{Spinoulas}\end{tabular} & 98.10                                                                    & 98.56                                                                   & 94.80                                                                        & 96.80                                                                      &97.11                        \\ \hline
Zhang et al. \cite{b31_zhang}                                                                                      & 97.01                                                                    & 97.81                                                                   & 95.42                                                                        & 97.02                                                                      & 96.82                        \\ \hline
Jung et al. \cite{b47_jung2}                                                                                       & 98.60                                                                    & 96.20                                                                   & 90.50                                                                        & 95.80                                                                      & 95.27                        \\ \hline
LivDet 2015 Winner \cite{b34_livdet2015}                                                                          & 98.10                                                                    & 95.40                                                                   & 93.72                                                                        & 94.36                                                                      & 95.39                        \\ \hline
Uliyan et al. \cite{b4_uliyan}                                                                                     & 95.00                                                                    & -                                                                       & -                                                                            & -                                                                          & 95.00                        \\ \hline

Kim et al. \cite{b23_kim}                                                                                         & -                                                                        & -                                                                       & -                                                                            & -                                                                          & 86.39                        \\ \hline
             { \textbf{MoSFPAD}}                                                              & { \textbf{98.39}}                                    & { \textbf{98.18}}                                   & { \textbf{95.35}}                                        & { \textbf{97.19}}                                      & { \textbf{97.23}} \\ \hline
\end{tabular}
}
\end{center}
\end{table*}
The comparison with state-of-the-art methods tested on LivDet 2017 database is mentioned in Table \ref{tab: Comparison_Intra_2017}. It shows that the performance of the proposed method is better than that of the methods discussed in \cite{b14_chugh1}, \cite{b15_chugh2}, \cite{b31_zhang}, and \cite{GONZALEZ-SOLER} on the datasets collected with digital persona and orcanthus sensors while it is comparable to them on greenbit dataset. This analysis concludes that the proposed method is capable of performing well while the spoof samples are fabricated using unknown materials. 
\begin{table}[!bht]
\begin{center}

\caption{Comparison with state-of-the-art methods on LivDet 2017 database in intra-sensor paradigm}
\label{tab: Comparison_Intra_2017}
\resizebox{0.5\textwidth}{!}{
\begin{tabular}{|p{2.5 cm} |c| c| c| c|}
\hline
\textbf{Method} & \textbf{\begin{tabular}[c]{@{}c@{}}Accuracy \\(Orcanthus)\end{tabular}}& \textbf{\begin{tabular}[c]{@{}c@{}}Accuracy \\(Digital Persona)\end{tabular}} & \textbf{\begin{tabular}[c]{@{}c@{}}Accuracy \\(Greenbit)\end{tabular}} & \multicolumn{1}{l|}{\textbf{Average}} \\ \hline
Chugh et al. \cite{b14_chugh1}              & 95.01                                           &   95.20                                                 & 97.42                                            & 95.88                                 \\ \hline
Chugh et al. \cite{b15_chugh2}              & 94.51                                            & 95.12                                                  & 96.68                                            & 95.43                                 \\ \hline
Zhang et al. \cite{b31_zhang}              & 93.93                                            & 92.89                                                  & 95.20                                            & 94.00                                 \\ \hline
Gonzalez et al. \cite{GONZALEZ-SOLER}              & 94.38                                            & 95.08                                                  & 94.54                                            & 94.66                                 \\ \hline
                 \textbf{MoSFPAD}           & \textbf{95.94}                                   & \textbf{95.40}                                         & \textbf{93.79}                                   & \textbf{95.05}                        \\ \hline
\end{tabular}
}
\end{center}
\end{table}

\begin{table}[!bht]
\begin{center}
\caption{Comparison with state-of-the-art methods on LivDet 2019 database in intra-sensor paradigm}
\label{tab: Comparison_Intra_2019}
\resizebox{0.5\textwidth}{!}{
\begin{tabular}{|p{2.5 cm} |c| c| c| c|}
\hline
\textbf{Method} & \textbf{\begin{tabular}[c]{@{}c@{}}Accuracy \\(Orcanthus)\end{tabular}}& \textbf{\begin{tabular}[c]{@{}c@{}}Accuracy \\(Digital Persona)\end{tabular}} & \textbf{\begin{tabular}[c]{@{}c@{}}Accuracy \\(Greenbit)\end{tabular}} & \multicolumn{1}{l|}{\textbf{Average}} \\ \hline

Jung CNN \cite{b30_orru}            & 99.13                                           &81.23                                                   &  99.06                                           & 93.14                                \\ \hline
Chugh et al. \cite{b15_chugh2}              &    97.50                                        &     83.64                                              &99.73                                             & 93.62                                \\ \hline
JWL LivDet \cite{b30_orru}              &      97.45                                      &       88.86                                            & 99.20                                            &   95.17                              \\ \hline
ZJUT Det A \cite{b30_orru}              &       97.50                                     &      88.77                                             &    99.20                                         &              95.16                   \\ \hline

                 \textbf{MoSFPAD}           & \textbf{97.72}                                   & \textbf{90.59}                                         & \textbf{97.31}                                   & \textbf{95.20}                        \\ \hline
\end{tabular}
}
\end{center}
\end{table}
Similarly, Table \ref{tab: Comparison_Intra_2019} reports a comparison of the proposed model's findings with state-of-the-art methods tested on LivDet 2019 database. It shows that the proposed method outperforms the method discussed in \cite{b15_chugh2} as well the FPAD algorithms i.e., JungCNN, JWL LivDet, ZJUT DET with an average classification accuracy of 95.20\%.
The comparative analysis of various LivDet databases indicates that the proposed method consistently performs better regardless of the sensors in the intra-sensor paradigm of FPAD whether the spoof samples are fabricated using known or unknown materials.
\subsubsection{\textbf{Comparison with Existing Approaches in Cross-Sensor Paradigm}}
The performance of the proposed model is also compared with state-of-the-art methods in the cross-sensor paradigm over LivDet 2011, 2013, and 2017 databases which are described in Table \ref{tab: comp_cross_2011}, Table \ref{tab: comp_cross_2013}, and Table \ref{tab: comp_cross_2017}, respectively. Table \ref{tab: comp_cross_2017} shows that the proposed method shows better PAD capability than the methods discussed in \cite{b14_chugh1}, and \cite{b31_zhang} over LivDet 2017 in the aforementioned scenario. Similarly, a comparison mentioned in Tables \ref{tab: comp_cross_2011}, and \ref{tab: comp_cross_2013} indicate the efficacy of the proposed method over the methods discussed in \cite{Nogueira} on LivDet 2011 and the methods discussed in \cite{b15_chugh2}, and \cite{Nogueira} on LivDet 2013 databases. \\ The comparison of the performance  concludes that the ensemble of SVC and the CNN architecture in the proposed manner improves the training of CNN in a better way. This hybrid architecture exhibits better performance than the state-of-the-art methods while being tested in the cross-sensor scenario.

\begin{table}[!bht]
\begin{center}
\caption{Comparison with state-of-the-art methods on LivDet 2011 database in cross-sensor paradigm} 
\resizebox{0.5\textwidth}{!}{
\begin{tabular}{|p{3 cm}|c|c|}
\hline
\textbf{\begin{tabular}[c]{@{}c@{}}Sensor \\ Training (Testing)\end{tabular}} & \textbf{Nogueira et al. \cite{Nogueira}} & {\textbf{\begin{tabular}[c]{@{}c@{}} MoSFPAD\end{tabular}}}  \\ \hline
Biometrika (Italdata)             & 62.8          & 57.28                                                                                                           \\ \hline
Italdata (Biometrika)         & 69          &  77.71                                                                                                           \\ \hline
\textbf{Average}                                                             & \textbf{65.9} & { \textbf{67.49}}                                                                                                   \\ \hline
\end{tabular}
}
\label{tab: comp_cross_2011}
\end{center}

\end{table}

\begin{table}[!bht]
\begin{center}
\caption{Comparison with state-of-the-art methods on LivDet 2013 database in cross-sensor paradigm} 
\resizebox{0.5\textwidth}{!}{
\begin{tabular}{|p{3.5 cm} |c| c| c|}
\hline
\textbf{\begin{tabular}[c]{@{}c@{}}Sensor \\ Training (Testing)\end{tabular}} & \textbf{Nogueira et al. \cite{Nogueira}} & \textbf{Chugh et al. \cite{b15_chugh2}} & {\textbf{\begin{tabular}[c]{@{}c@{}}MoSFPAD\end{tabular}}} \\ \hline
Biometrika (Italdata) &   91.2                                                    & 95.7          & 94.07                                                                                                           \\ \hline
Italdata (Biometrika)                                           & 97.7            & 96.50          &  99.30                                                                                                           \\ \hline

\textbf{Average}                                                             & \textbf{94.45}   & \textbf{96.1} & { \textbf{96.68}}                                                                                                   \\ \hline
\end{tabular}}

\label{tab: comp_cross_2013}
\end{center}
\end{table}

\begin{table}[!hbt]
\begin{center}
\caption{Comparison with state-of-the-art methods on LivDet 2017 database in cross-sensor paradigm} 
\label{tab: comp_cross_2017}
\resizebox{0.5\textwidth}{!}{
\begin{tabular}{|p{3 cm}|c|c|c|}
\hline
\textbf{\begin{tabular}[c]{@{}c@{}}Sensor \\ Training (Testing)\end{tabular}} & \textbf{Zhang et al. \cite{b31_zhang}} & \textbf{Chugh et al. \cite{b15_chugh2}} & { \textbf{\begin{tabular}[c]{@{}c@{}} MoSFPAD\end{tabular}}}  \\ \hline
GreenBit (Orcanthus)                                                 & 43.98            & 49.43          & 57.33                                                                                                           \\ \hline
GreenBit (Digital Persona)                                           & 80.39            & 89.37          &  83.55                                                                                                           \\ \hline
Orcanthus (GreenBit)                                                 & 68.82            & 69.93          & 72.58                                                                                                           \\ \hline
Orcanthus (Digital Persona)                                          & 62.30            & 57.99          & 61.29                                                                                                           \\ \hline
Digital Persona (GreenBit)                                           & 87.9             & 89.54          & 77.39                                                                                                            \\ \hline
Digital Persona (Orcanthus)                                          & 44.30            & 49.32          & 58.25                                                                                                           \\ \hline
\textbf{Average}                                                             & \textbf{64.61}   & \textbf{67.59} & { \textbf{68.39}}                                                                                                   \\ \hline
\end{tabular}}

\end{center}

\end{table}

\subsubsection{\textbf{Evaluation of MoSFPAD in High-Security Systems}\label{evaluation in high-security}}
\begin{figure*}[!h]
	\centering
	\resizebox{0.7\textwidth}{!}{
	
	\subfigure
	{	
		\includegraphics[width=0.50\textwidth]{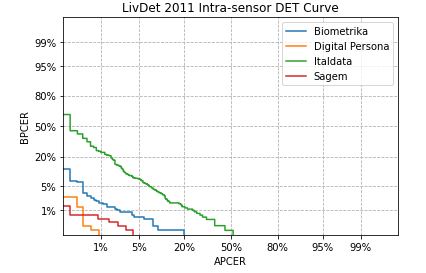}
		\hspace{4 mm}
		\includegraphics[width=0.50\textwidth]{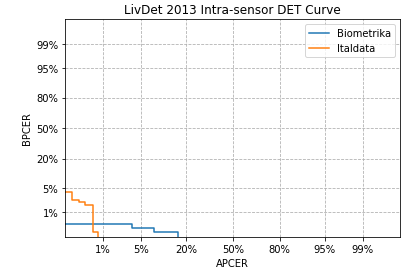}
	}}
	\resizebox{0.7\textwidth}{!}{
	\subfigure
	{
		\includegraphics[width=0.50\textwidth]{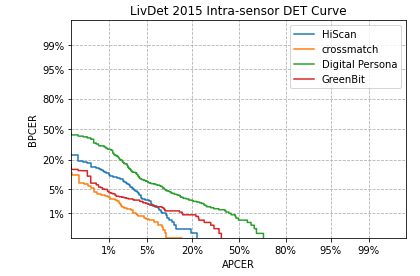}
		\hspace{4 mm}
		\includegraphics[width=0.50\textwidth]{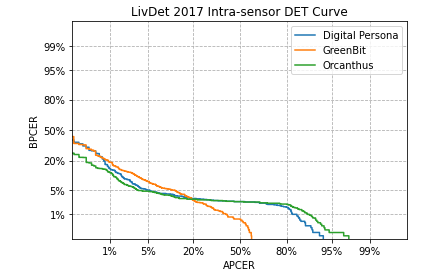}
	}}
	\resizebox{0.35\textwidth}{!}{
	
	\subfigure
	{	
		\includegraphics[width=0.50\textwidth]{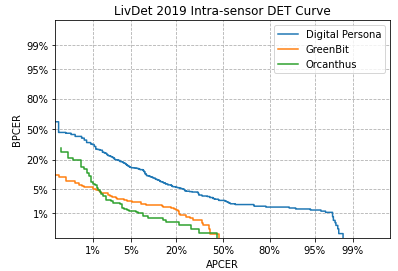}
		\hspace{4 mm}
	}}
	\caption{Detection Error Trade-off (DET) curves for LivDet 2011, 2013, 2015, 2017, and 2019 databases}
	\label{DET_Cur}
	\vspace{-2mm}
	
\end{figure*}

The main goal of any FPAD system is not only to perform in a single perspective either to gain low APCER or BPCER but also to perform in more realistic conditions. The model is required to be tested for high-security performance also and for the same we have utilized the DET graph. A DET graph is a graphical plot of error rates for binary classification systems, plotting the APCER versus BPCER. In this section, We have reported the DET curves for all the databases of LivDet 2011, 2013, 2015, 2017 and 2019, which are depicted in Fig. \ref{DET_Cur}.
It can be observed that the model shows consistent performance for LivDet 2013, and 2017 databases as the curves plotted for their sensors data are similar and close to each other. We also have analyzed the performance of the proposed model in a high-security environment by referring to these DET curves. For a high-security system, a low APCER is desired. It can be observed that for LivDet 2011,  for 1\% of APCER, BPCER ranges from 0.2\% to 2\% for biometrika, sagem, and digital persona while it is 22\% for italdata sensor. For LivDet 2013, the model exhibits consistent performance as the BPCER varies in a small range, and for both of the sensors, the BPCER is less than 1\%. The impact on the performance in the cross-material paradigm is clearly visible for the curves on LivDet 2015, 2017, and 2019 databases. In LivDet 2015, the BPCER is less than 5\% for crossmatch and greenbit sensors while it is 10\% and 30\% for the remaining two sensors. The curves for the sensors of LivDet 2017 are close to each other and the BPCER is less than 20\% for 1\% of APCER. For LivDet 2019, the model shows good performance on greenbit and orcanthus sensors as the BPCER is less than 7\% while it is 30\% for digital persona.
\subsubsection{\textbf{Processing Time}}
The processing time of an FPAD model is considered the amount of time it takes to find whether the input fingerprint sample is live or spoof. This time is supposed to be minimum as the sample has to undergo the process of verification after the detection of its liveness. The suggested model, $MoSFPAD$, takes the classification time of 95 milli-seconds on Intel(R) Core(TM) i5-6500 CPU @ 3.20GHz $6^{th}$ generation processor to classify a single fingerprint image. This achievement makes it suitable for devices that have a processor with minimal computational power.
\section{Conclusion}
\label{conclusion}
In this paper, we have proposed a novel end-to-end architecture that utilizes MobileNet as a feature extractor and SVC as a classifier. The involvement of the SVC at the top enables the MobileNet to learn its parameters in a better way than the sigmoid or soft-max activation functions. The suggested model is capable to perform PAD in intra-sensor cross-material and cross-sensor paradigms with better classification capability as compared with state-of-the-art methods. The proposed hybrid model is the first of its kind as best of our knowledge. The end-to-end collaboration of both tools also makes it efficient in terms of processing time and suitable for various real-time applications required to ensure the security of fingerprint-based recognition systems.

\bibliographystyle{IEEE}
\bibliography{references}
\vspace{-1 cm}
\begin{IEEEbiography}[{\includegraphics[width=1in,height=1.25in,clip,keepaspectratio]{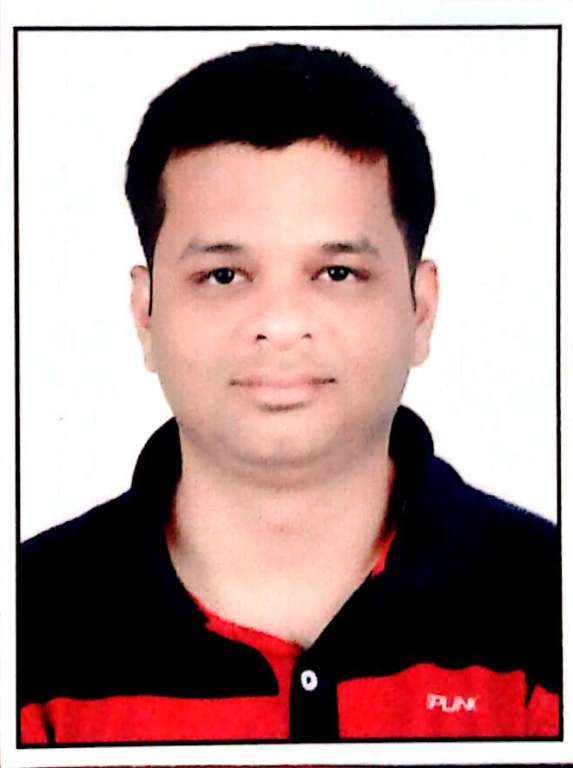}}]{Anuj Rai} received his M.Tech. degree in Computer Technology and Applications from National Institute of Technical Teachers Training and Research Bhopal, India, in 2015. He is currently pursuing his Ph.D. in Computer Science and Engineering department at Indian Institute of Technology Indore.
\end{IEEEbiography}
\vspace{-1 cm}
\begin{IEEEbiography}[{\includegraphics[width=1in,height=1.25in,clip,keepaspectratio]{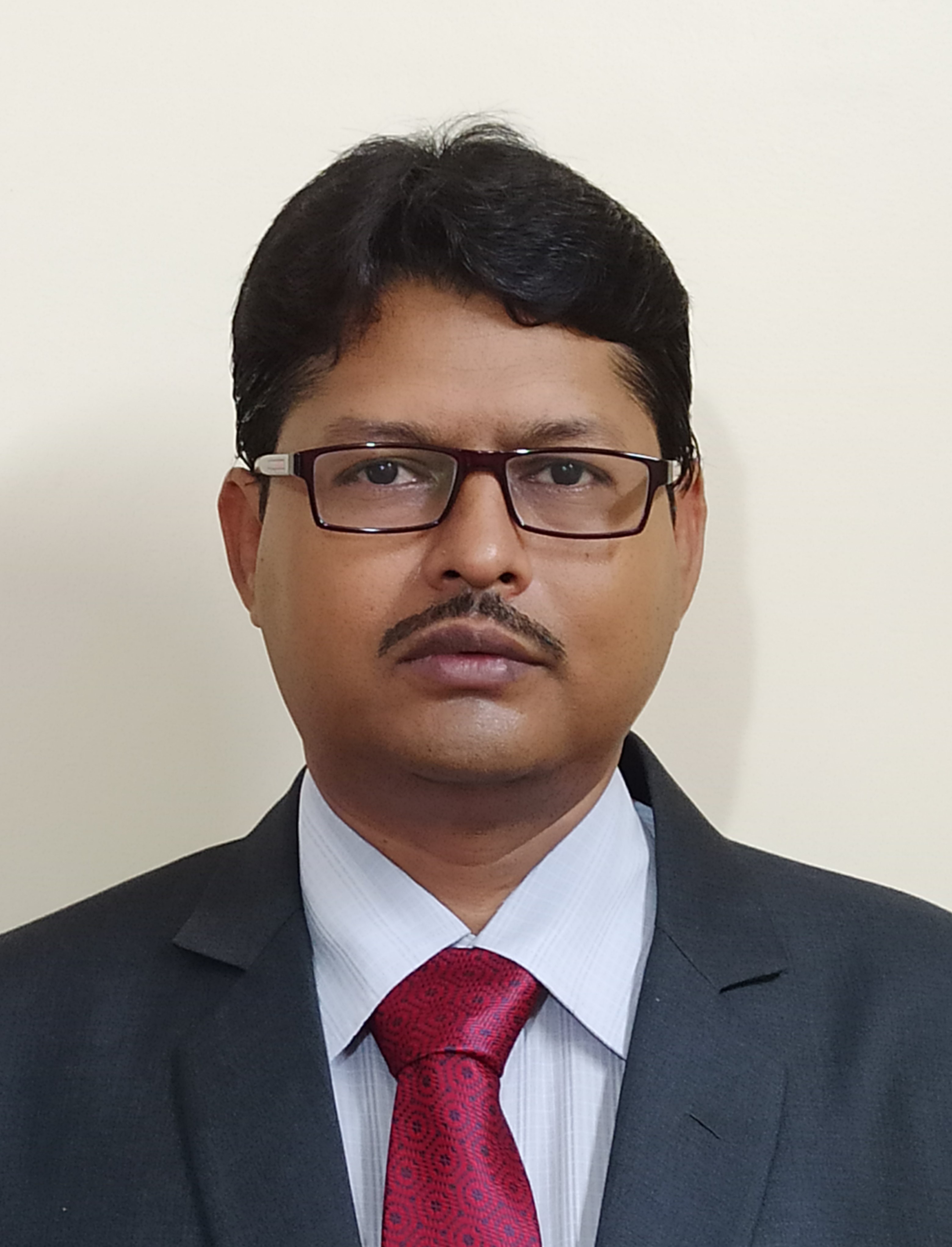}}]{Somnath Dey} is currently working as an Associate Professor in the Discipline of Computer Science \& Engineering at the Indian Institute of Technology Indore (IIT Indore). He received his B. Tech. degree in Information Technology from the University of Kalyani in 2004. He completed his M.S. (by research) and Ph.D. degree in Information Technology from the School of Information Technology, Indian Institute of Technology Kharagpur, in 2008 and 2013, respectively. His research interest includes biometric security, biometric template protection, biometric cryptosystem, and traffic sign detection.
\end{IEEEbiography}
\vspace{-0.5 cm}
\begin{IEEEbiographynophoto}{Pradeep Patidar} 
Completed his Bachelor of Technology (B.Tech.) degree from the department of Computer Science of Engineering, Indian Institute of Technology Indore in 2022.
\end{IEEEbiographynophoto}
\vspace{-10mm}
\begin{IEEEbiographynophoto}{Prakhar Rai} 
Completed his Bachelor of Technology (B.Tech.) degree from the department of Computer Science of Engineering, Indian Institute of Technology Indore in 2022.
\end{IEEEbiographynophoto}

\end{document}